\documentclass[journal,compsoc]{IEEEtran}
\usepackage[noadjust]{cite}

\usepackage[group-separator={,}]{siunitx}

\usepackage{graphicx}

\usepackage{amsmath}
\usepackage{amssymb}
\usepackage{bm}
\usepackage{subcaption} 

\ifCLASSINFOpdf
\else
\fi

\begin{document}
%
\title{Neural Image Compression for \\Gigapixel Histopathology Image Analysis}
%
%
%

\author{David~Tellez*, 	
Geert Litjens, Jeroen~van~der~Laak, Francesco~Ciompi
\thanks{*D. Tellez, 	
G. Litjens, J. van der Laak, and F. Ciompi are with the Diagnostic Image Analysis Group and the Department of Pathology, Radboud University Medical Center, 6500HB Nijmegen, The Netherlands (corresponding e-mail: david.tellezmartin@radboudumc.nl).}
}




\IEEEtitleabstractindextext{%
\begin{abstract}
We propose Neural Image Compression (NIC), a two-step method to build convolutional neural networks for gigapixel image analysis solely using weak image-level labels. First, gigapixel images are compressed using a neural network trained in an unsupervised fashion, retaining high-level information while suppressing pixel-level noise. Second, a convolutional neural network (CNN) is trained on these compressed image representations to predict image-level labels, avoiding the need for fine-grained manual annotations. We compared several encoding strategies, namely reconstruction error minimization, contrastive training and adversarial feature learning, and evaluated NIC on a synthetic task and two public histopathology datasets. We found that NIC can exploit visual cues associated with image-level labels successfully, integrating both global and local visual information. Furthermore, we visualized the regions of the input gigapixel images where the CNN attended to, and confirmed that they overlapped with annotations from human experts.
\end{abstract}

\begin{IEEEkeywords}
Gigapixel image analysis, computational pathology, convolutional neural networks, representation learning.
\end{IEEEkeywords}
}

\maketitle

\IEEEdisplaynontitleabstractindextext

%
\IEEEpeerreviewmaketitle

\section{Introduction}
\label{sec:introduction}

\textsc{Gigapixel} images are three-dimensional arrays composed of more than 1 billion pixels; these are common in fields like Computational Pathology~\cite{LITJENS201760} and Remote Sensing~\cite{zhang2016deep}{}, and are often associated with labels at image level. The fundamental challenge of gigapixel image analysis with weak image-level labels resides in the low signal-to-noise ratio present in these images. Typically, the signal consists of a subtle combination of high- and low-level patterns that are related to the image-level label, while most of the pixels behave as distracting noise. Furthermore, the nature and spatial distribution of the signal are both unknown, often referred to as the \emph{what} and the \emph{where} problems, respectively. 

\subsection{The \emph{what} and the \emph{where} problems}

Researchers have addressed the challenge of gigapixel image analysis by making different assumptions about the signal, simplifying either the \emph{what} or the \emph{where} problem. 

The most widespread simplification assumes that the signal is fully recognizable at a low level of abstraction, i.e., the image-level label has a patch-level representation. This simplification addresses the \emph{what} problem by decomposing the gigapixel image into a set of patches that can be independently annotated. Typically, these patches are manually annotated to perform automatic detection or segmentation using a neural network, relegating the task of performing image-level prediction to a rule-based decision model about the patch-level predictions~\cite{veta2018predicting,camelyon16,sirinukunwattana2017gland,LITJENS201760}. This assumption is not valid for image-level labels that do not have a known patch-level representation. Furthermore, patch-level annotation in gigapixel images is a tedious, time consuming and error-prone process, and limits what machine learning models can learn to the knowledge of human annotators.

Other researchers have assumed that the signal can exist at a low level of abstraction, but it is then not fully recognizable, i.e., the image-level label has a patch-level representation that is unknown to human annotators. Furthermore, the mere presence of these patches is enough evidence to make a prediction at the image level, ignoring the spatial arrangement between patches, thus solving the \emph{where} problem. Making this assumption falls into the multiple-instance learning (MIL) framework, which reduces the gigapixel image analysis problem into detecting patches that contain the true signal while suppressing the noisy ones~\cite{wang2018weakly,ilse2018attention,combalia2018monte,tomczak2018histopathological, hou2016patch}. However, these methods can only take into account patterns present within individual patches, neglecting the potential relationships among them. More generally, MIL techniques cannot exploit patterns present in higher levels of abstraction since they ignore the spatial distribution among patches. This is also true for methods that aggregate patch-level information by means of spatial pooling~\cite{wang2018weakly,coudray2018classification}.

In this work, we do not make any assumptions about the nature or spatial distribution of the visual cues associated with image-level labels. We argue that convolutional neural networks (CNN) are designed to solve the \emph{what} and the \emph{where} problems simultaneously~\cite{Goodfellow-et-al-2016}, and propose a method to use them for gigapixel image analysis. However, feeding CNNs directly with gigapixel images is computationally unfeasible. Instead, we propose Neural Image Compression (NIC), a technique that maps images from a low-level pixel space to a higher-level latent space using neural networks. In this way, gigapixel images are compressed into a highly compact representation, which can be used to train a CNN using a single GPU for predicting any kind of image-level label.

\begin{figure*} [t]
\begin{center}
\includegraphics[width=1\textwidth]{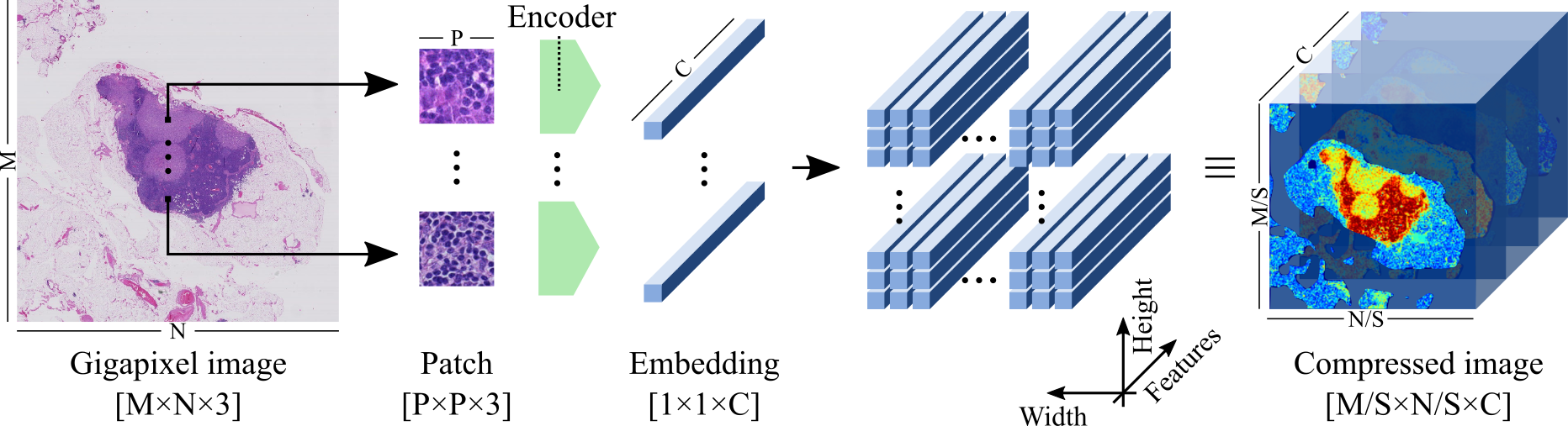}
\end{center}
\caption[pipeline] 
{ \label{fig:pipeline} Gigapixel neural image compression. Left: a gigapixel histopathology whole-slide image is divided into a set of patches mapped to a set of low-dimensional embedding vectors using a neural network (the encoder). Center: these embeddings are stored keeping the spatial arrangement of the original patches. Right: the resulting array is a compressed representation of the gigapixel image. $M$ and $N$: size of the gigapixel image; $P$: size of the square patches; $C$: size of the embedding vectors; and $S$: stride used to sample the patches. Typically: $M=N=\num{50000}$ and $P=S=C=128$.}
\end{figure*} 

\subsection{Neural Image Compression}

Gigapixel NIC was designed to reduce the size of a gigapixel image while retaining semantic information by shrinking its spatial dimensions and growing along the feature direction (see Fig.~\ref{fig:pipeline}). The method works by, first, dividing the gigapixel image into a set of high-resolution patches. Second, each high-resolution patch is compressed with a neural network (the \emph{encoder}) that maps every image into a low-dimensional embedding vector. Finally, each embedding is placed into an array that keeps the original spatial arrangement intact so that neighbor embeddings in the array represent neighbor patches in the original image. 

NIC was inspired by cognitive mechanisms. Human observers can describe complex visual patterns using only a few words without needing to describe each individual pixel. Similarly, the \emph{encoder} can describe patches with low-dimensional embedding vectors, ignoring superfluous details. It is a powerful method that competes with classical approaches in terms of compression rate~\cite{theis2017lossy}. Moreover, previous works on representation learning and transfer learning have demonstrated that neural networks excel at extracting features that can be exploited by other networks to solve a variety of downstream tasks~\cite{donahue2016adversarial,dumoulin2016adversarially,oord2018representation,van2017neural}. This makes NIC an ideal candidate for reducing the size of gigapixel images before feeding a CNN.

The \textit{encoder} network can be trained using a wide variety of techniques. In this work, we selected and compared representative methods from three well-known families of unsupervised representation learning algorithms: reconstruction error minimization, contrastive training, and adversarial feature learning. First, autoencoders (AE) have been proposed as a straightforward method to learn a compact representation of a given data manifold~\cite{Goodfellow-et-al-2016}. AEs are neural networks that follow a particular encoder-bottleneck-decoder architecture. They aim to reconstruct input images by minimizing a reconstruction loss, e.g., the mean squared error (MSE). In particular, we considered the case of the variational autoencoder (VAE), a powerful modification of the original AE that relies on a probabilistic approach~\cite{kingma2013auto}. Second, we investigated a discriminative model based on contrastive training~\cite{oord2018representation,koch2015siamese,melekhov2016siamese,hyvarinen2016unsupervised}. This model senses the world via an encoding network that maps images to embedding vectors. By training this model to distinguish between pairs of images with \emph{same} or \emph{different} semantic information, the encoder is enforced to learn a compact representation of the input data. Third, we investigated adversarial feature learning~\cite{donahue2016adversarial,dumoulin2016adversarially}, a training framework based on Generative Adversarial Networks (GAN)~\cite{NIPS2014_5423}. GANs emerged as powerful generative models that map low-dimensional latent distributions into complex data. There is evidence that these latent spaces capture some of the high-level semantic information present in the data~\cite{chen2016infogan}. However, standard GAN models do not support the reverse operation, i.e., mapping data to the latent space. The Bidirectional GAN model (BiGAN~\cite{donahue2016adversarial}) learns this mapping using an explicit encoding network in the training loop. Intuitively, the encoder benefits from all the high-level features which were fully automatically discovered by the generator.

\subsection{Gigapixel Image Analysis}

Without any loss of generality, we applied our method to two of the largest publicly available histopathology datasets to demonstrate its effectiveness in real-world applications: the \textit{Camelyon16} Challenge~\cite{camelyon16} and the \textit{TUPAC16} Challenge~\cite{veta2018predicting}. These datasets consist of gigapixel images of human tissue acquired with brightfield microscopy at very high magnification, also known as whole-slide images (WSI). These WSIs were stained with hematoxylin and eosin (H\&E), the most widely used stain in routine histopathology diagnostics, that highlights general tissue morphology such as cell nuclei and cytoplasm. Each WSI is associated with a single image-level label: the presence of tumor metastasis for \textit{Camelyon16}, and the tumor proliferation speed based on gene-expression profiling for \textit{TUPAC16}. 

A benefit of using a CNN for gigapixel image analysis is that, once trained, the CNN's areas of interest in the input image can be visualized using gradient-weighted class-activation maps (Grad-CAM)~\cite{selvaraju2017grad}{}. These saliency maps provide an answer to the \emph{where} problem by locating visual cues related to the image-level labels. Identifying visual evidence for CNN predictions is of utmost importance in the medical domain regarding algorithm interpretation and knowledge discovery. For the first time, we performed this saliency analysis on gigapixel images and compared the resulting maps with the patch-level annotations of an expert observer.

\subsection{Contributions}

This work is an extension of our conference paper~\cite{Tell18b}. A number of additions have been made: three new datasets, an additional encoding method, the Grad-CAM analysis, a new experiment at the patch level, a new experiment at the image level, a more thorough evaluation using cross-validation, and an independent test evaluation performed by a third-party.

Our contributions can be summarized as follows:

\begin{itemize}
\item We propose Neural Image Compression (NIC) as a method to reduce gigapixel images to highly-compact representations, suitable for training a CNN end-to-end to predict image-level labels using a single GPU and standard deep learning techniques. 
\item We compared several encoding methods that map high-resolution image patches to low-dimensional embedding vectors based on different unsupervised learning techniques: reconstruction error minimization, contrastive training, and adversarial feature learning. 
\item We evaluated NIC in three publicly available datasets: a synthetic set designed to evaluate the method; and two histopathological breast cancer sets of whole-slide images used to train the system to predict the presence of tumor metastasis and the tumor proliferation speed. 
\item We generated saliency maps representing the CNN's areas of interest in the image in order to discover and localize visual cues associated to the image-level labels.
\end{itemize}

The paper is organized as follows: Sec. \ref{sec:wsi_compression} and Sec. \ref{sec:wsi_classification} describe the methods in depth; Materials and experimental results are described in Sec. \ref{sec:experiments_results}; the discussions and conclusions are stated in Sec. \ref{sec:discussion} and Sec. \ref{sec:conclusion}, respectively.

\section{Neural Image Compression}
\label{sec:wsi_compression}

Let us define $\omega \in \mathbb{R}^{M\times N\times 3}$ as the gigapixel image (e.g., a WSI) to be compressed, with $M$ rows, $N$ columns, and three color channels (RGB). In order to compress $\omega$ into a more compact representation $\omega'$, two steps were taken. First, $\omega$ was divided into a set of high-resolution patches $X = \{ x_{ij}\}$ with $ x_{ij} \in \mathbb{R}^{P\times P\times 3}$, sampled from the $i$-th row and $j$-th column of an uniform grid of square patches of size $P$ using a stride of $S$ throughout $\omega$. Second, each $x_{ij}$ was compressed independently from each other, generating a set of low-dimensional embedding vectors of length $C$ at each spatial location on the grid: $Y = \{ e_{ij}\}$ with $e_{ij} \in \mathbb{R}^{C}$.

We formulated the task of mapping high-entropy $X$ into low-entropy $Y$ as an instance of an unsupervised representation learning problem, and parameterized this mapping function with a neural network $E$ so that $X \xrightarrow{E}Y$. By sliding $E$ throughout all $ij$ spatial locations, $\omega$ was compressed into $\omega'$ with a total volume reduction of $F = 3\frac{S^2}{C}$. More formally:

\begin{equation}
\label{eq:wsi_compression}
\omega \in \mathbb{R}^{M\times N\times 3} \xrightarrow{E} \omega' \in \mathbb{R}^{\frac{M}{S} \times \frac{N}{S} \times C}
\end{equation}

We investigated several unsupervised encoding strategies for learning $E$. Three of the most well-known and accessible methods in unsupervised image representation learning were selected. In all cases, neural networks were trained to solve an auxiliary task and learn $E$ as a by-product of the training process. Note that none of the studied methods required the use of manual annotations. Network architectures and training protocols are detailed in the Supplementary Material accompanying this paper.

\begin{figure} [t]
\centering
\includegraphics[width=0.4\textwidth]{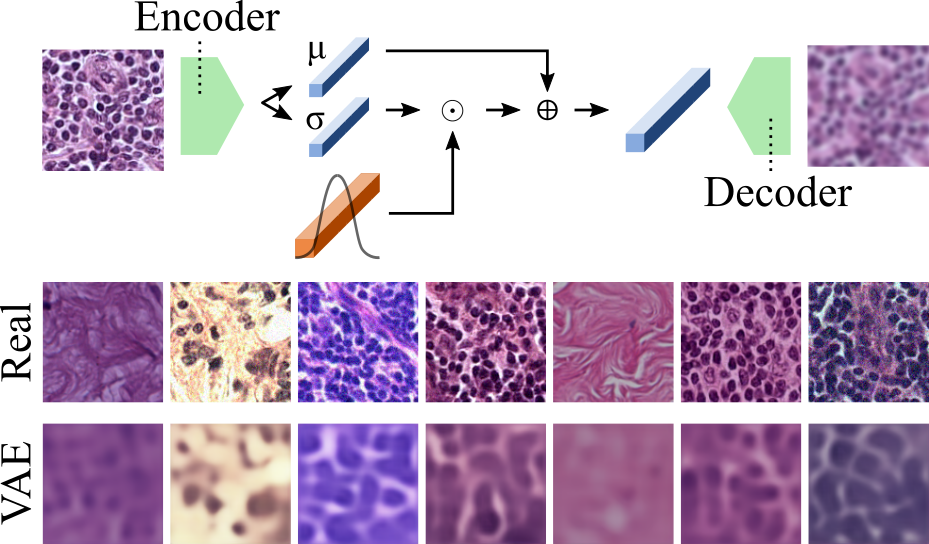}
\caption[vae] 
{ \label{fig:vae} Variational Autoencoder. Top: the encoder maps a patch to an embedding vector depending on a noise vector while the decoder reconstructs the original patch from the embedding vector. Bottom:
pairs of real and reconstructed patch samples using $C=128$.}
\end{figure}

\subsection{Variational Autoencoder}

Two networks are trained simultaneously, the encoder $E$ and the decoder $D$. The task of $E$ is to map an input patch $x$ into a compact embedded representation $e$, and the task of $D$ is to reconstruct $x$ from $e$, producing $x'$. In this work, we used a more sophisticated version of AE, the variational autoencoder (VAE)~\cite{kingma2013auto}. The encoder in the VAE model learns to describe $x$ with an entire probability distribution instead of a single vector (Fig.~\ref{fig:vae}). More formally, $E$ outputs $\mu\in~\mathbb{R}^{C}$ and $\sigma\in~\mathbb{R}^{C}$, two embeddings representing the mean and standard deviation of a normal distribution such that:

\begin{equation}
\label{eq:vae}
e = \mu + \sigma \odot n
\end{equation}

with $n \sim \mathcal{N}(0, 1)$ and $\odot$ denoting element-wise multiplication. 

We trained the VAE model by optimizing the following objective:

\begin{multline}
\label{eq:vae_objective}
\mathcal{V}_{\text{VAE}}(x, n, \theta_E, \theta_D) = \\
= \underset{E, D}{\min{}} \Big[ \underbrace{ \big(x - D(E(x, n))\big)^2}_\text{Reconstruction error} + \underbrace{ \gamma (1 + \log{\sigma^2} - \mu^2 - \sigma^2)}_\text{KL divergence} \Big]
\end{multline}

with $\gamma$ as a scaling factor, and $\theta_E$ and $\theta_D$ as the parameters of $E$ and $D$, respectively. Note that we optimized $\theta_E$ and $\theta_D$ to minimize both the reconstruction error between the input and output data distributions, and the KL divergence between the embedding distribution and the normal $\mathcal{N}(0, 1)$ distribution.

This procedure results in a continuous latent space where changes in the embedding vectors are proportional to changes in the input data and vice-versa, effectively retaining semantic knowledge present in the input space.

\begin{figure} [t]
\centering
\includegraphics[width=0.4\textwidth]{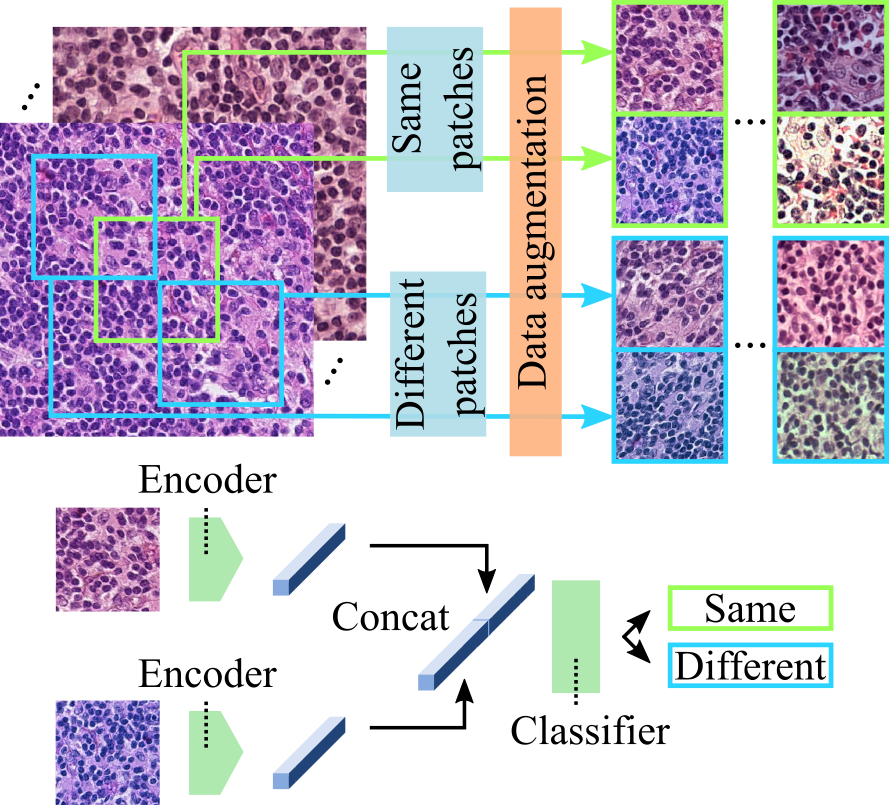}
\caption[contrastive] 
{ \label{fig:contrastive} Contrastive training. Top: pairs of patches are extracted from gigapixel images. Pairs labeled as \emph{same} originate from the same spatial location whereas \emph{different} are extracted from either adjacent locations or different images. Bottom: scheme of a Siamese network trained for binary classification using the previous pairs.}
\end{figure}

\subsection{Contrastive Training}

We assembled a training dataset composed of pairs of patches $\boldsymbol{x} = \{x^{(1)}, x^{(2)}\}$ where each pair $\boldsymbol{x}$ was associated with a binary label $y$. Each label described whether the patches had been extracted from the \emph{same} or a \emph{different} location in a given gigapixel image, with $y=1$ and $y=0$, respectively. We trained a two-branch Siamese network~\cite{melekhov2016siamese} to solve this classification problem (Fig.~\ref{fig:contrastive}). 

We applied heavy data augmentation on all patches as indicated in~\cite{tellez2018whole}, i.e., rotation, color augmentation, brightness, contrast, zooming, elastic deformation, and added Gaussian noise. Due to the strong augmentation, patches from the \emph{same} location looked substantially different in a highly non-linear fashion while keeping a similar overall structure (semantic), see examples in Fig.~\ref{fig:contrastive}. Patches from the \textit{different} class were extracted from two distributions: 75\% of them corresponded to non-overlapping adjacent locations (i.e., neighboring patches) where most of the visual features were shared, and the remaining 25\% were sampled from different WSIs. Note that we included more of the neighboring \textit{different} pairs to increase the difficulty of the classification task, forcing the network to extract higher-level features. The same data augmentation was applied to other encoders as well to ensure a fair comparison.

\begin{figure} [t]
\centering
\includegraphics[width=0.4\textwidth]{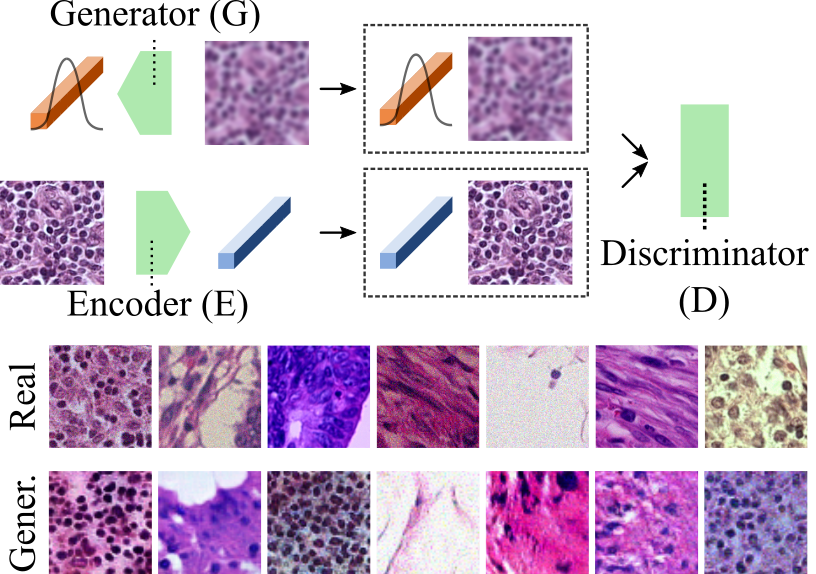}
\caption[bigan] 
{ \label{fig:bigan} Adversarial Feature Learning. Top: three networks play a minimax game where the discriminator distinguishes between \emph{actual} or \emph{generated} image-embedding pairs, while the generator and the encoder fool the discriminator by producing increasingly more realistic images and embeddings. Bottom: real and generated patch samples using $C=128$.
}
\end{figure} 

\subsection{Bidirectional Generative Adversarial Network}

The BiGAN setup consists of three networks: a generator $G$, a discriminator $D$, and an encoder $E$ (Fig~\ref{fig:bigan}). $G$ maps a latent variable $z~\sim \mathcal{N}(0, 1)$ to generated images $x'$:

\begin{equation}
\label{eq:bigan_g}
z~\sim \mathcal{N}(0, 1) \in~\mathbb{R}^{C}  \xrightarrow{G} x'~\in~\mathbb{R}^{P\times P\times 3}
\end{equation}

whereas $E$ maps images $x$ sampled from the true data distribution $\mathcal{X}$ to embeddings $e$:

\begin{equation}
\label{eq:bigan_e}
x~\sim \mathcal{X} ~\in~\mathbb{R}^{P\times P\times 3} \xrightarrow{E} e \in~\mathbb{R}^{C} 
\end{equation}

During training, the three networks play a minimax game where the discriminator $D$ tries to distinguish between \emph{actual} or \emph{generated} image-embedding pairs, i.e., $\{x, e\}$ and $\{x', z\}$ respectively, while $G$ and $E$ try to fool $D$ by producing increasingly more realistic images $x'$ and embeddings $e$ closer to $\mathcal{N}(0, 1)$. More formally, we optimized the following objective function:

\begin{multline}
\label{eq:bigan_objective}
\mathcal{V}_{\text{BiGAN}}(x, z, \theta_G, \theta_E, \theta_D) = \\
=\underset{G, E}{\min{}}\underset{D}{\max{}} \Big[\log{\big[ D\big(x, \underbrace{E(x)}_{e}\big) \big]} + \log{\big[ 1 - D\big(\underbrace{G(z)}_{x'},z \big) \big]} \Big]
\end{multline}

with $\theta_G$, $\theta_E$, and $\theta_D$ representing the parameters of $G$, $E$, and $D$, respectively. 

The authors of BiGAN theoretically and experimentally demonstrate that $G$ and $E$ learn an approximate inverse mapping function from each other, producing an encoding network $E$ that learns a powerful low-dimensional representation of the image world inherited from $G$, suitable for downstream tasks such as supervised classification~\cite{donahue2016adversarial}.

\section{Gigapixel Image Analysis}
\label{sec:wsi_classification}

In this section, we describe a method to train a CNN to predict image-level labels directly from compressed gigapixel images. Furthermore, we analyzed the location of visual cues associated with the image-level labels.

\subsection{Feeding a CNN with compressed gigapixel images}

We consider a dataset of gigapixel images $\Omega = \{\omega_i \}_{i=1}^{Q}$ that were compressed into $\Omega' = \{\omega_i' \}_{i=1}^{Q}$ with $\omega_i' \in \mathbb{R}^{\frac{M}{S} \times \frac{N}{S} \times C}$ using Eq.~\ref{eq:wsi_compression}. In order to train a standard CNN on a dataset like $\Omega'$, we set the depth of the convolutional filters of the input layer to be equal to the code size $C$ used to compress the images.

We hypothesized that such a CNN can learn to detect highly discriminative features by exploiting two complementary sources of information from $\Omega'$: (1) the \emph{global} context encoded within the spatial arrangement of embedding vectors, and (2) the \emph{local} high-resolution information encoded within the features of each embedding vector. 

\subsection{Preventing overfitting}

Note that in this setup, despite its gigapixel nature, each compressed image $\omega_i'$ constitutes a single training data point. Most public datasets with gigapixel images and their respective image-level labels consist only of a few hundred data points~\cite{camelyon16,veta2018predicting}, increasing the risk of overfitting. The steps taken to prevent this effect are enumerated below. 

First, we extended the training dataset $\Omega'$ by taking spatial crops of size $R\times R \times C$ from $\omega_i'$, drastically increasing the total number and variability of the samples presented to the CNN~\cite{krizhevsky2012imagenet}. During training, we randomly sampled the location of the center pixel of these crops. During testing, we selected $T$ crops uniformly distributed along the spatial dimensions of $\omega_i'$ and averaged the predictions of the CNN across them~\cite{krizhevsky2012imagenet}. Without any loss of generality, we applied this method to histopathology WSIs. As WSIs often contain large empty areas with no tissue, we detected the tissue regions~\cite{bandi2017comparison} and sampled crops proportionally to the distance to background to accelerate the training, so that areas with higher tissue density were sampled more often. Similarly, test crops were sampled from locations where tissue was present.

The second measure taken to prevent overfitting was a simple augmentation at image level (i.e., 90-degree rotation and mirroring), encoding each image 8 times. This augmentation was carried out during testing as well, averaging the predictions of the CNN across them.

Finally, we designed a CNN architecture aimed at reducing the number of parameters present in the model. In particular, all convolutional layers were set to use depthwise separable convolutions, a type of convolution that reduces the number of parameters while maintaining a similar level of performance~\cite{chollet2016xception}.

\subsection{Visualizing visual cues related to image-level labels}

The problem of feature localization is of utmost relevance for gigapixel image analysis: visual cues related to the image-level labels are often sparse and positioned in arbitrary locations within the image. For the purpose of identifying the location of these visual cues, we applied the Gradient-weighted Class-Activation Map (Grad-CAM) algorithm~\cite{selvaraju2017grad}{} to our trained CNN. 

Given a compressed gigapixel image $\omega'$, its associated image-level label $y$, and a trained CNN, Grad-CAM performs a forward pass over $\omega'$ to produce a set of $J$ intermediate three-dimensional feature volumes $f_j^{(k)}$, with $j$ and $k$ indicating the $j$-th and $k$-th convolutional layer and feature map, respectively. Subsequently, it computes the gradients of $f_j^{(k)}$ with respect to $y$ for a fixed convolutional layer. It averages the gradients across the spatial dimensions and obtains a set of gradient coefficients $\gamma_j^{(k)}$, indicating how relevant each feature map is for the desired output $y$. Finally, it performs a weighted sum of the feature maps $f_j^{(k)}$ using the gradient coefficients $\gamma_j^{(k)}$:

\begin{equation}
\label{eq:bigan_e}
h^{(k)} = \sum_{j=1}^{J} f_j^{(k)} \gamma_j^{(k)}
\end{equation}

We applied the visualization method to the first convolutional layer ($k=1$) in order to maximize the heatmap resolution.

\begin{figure} [t]
\centering
\includegraphics[width=0.45\textwidth]{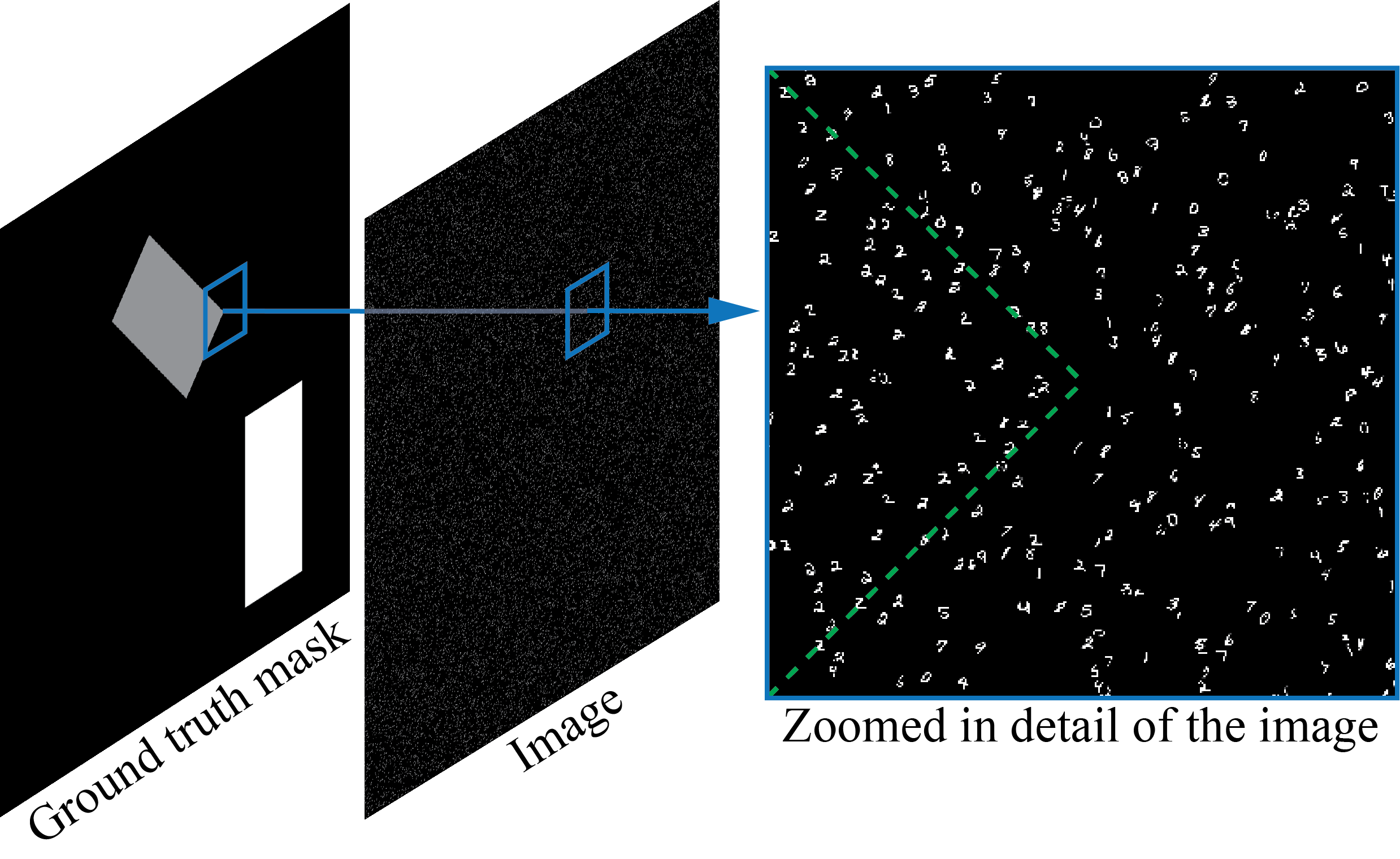}
\caption[nicmnisttask] 
{ \label{fig:synthetic_data} Example of an image from the synthetic dataset. Left: ground truth mask depicting the \textit{tilted} and \textit{non-tilted} rectangles that simulate lesions in grey and white, respectively. Center: image containing instances of MNIST digits; classes are defined by the rectangles or selected randomly. Right: all digits within the \textit{tilted} rectangle boundary (in green) belong to the same class (number two), which corresponds to the image label as well.
}
\end{figure}

\section{Experiments and Results}
\label{sec:experiments_results}

We conducted a series of experiments to evaluate the performance of gigapixel NIC. First, we evaluated NIC in synthetic data to gain an understanding of the method and how its hyper-parameters affect performance. Second, we applied the method to several public histopathological datasets.

\subsection{Materials}
\label{sec:materials}

In this work, a synthetic dataset and three histopathology cohorts from different sources were used for supervised and unsupervised training at patch and image level; patients and WSIs were unique across all cohorts.

\subsubsection{Synthetic dataset}
We developed and tested NIC with a synthetic dataset that mimicked the task of end-to-end WSI analysis before deploying it with real WSIs. As a substitute for WSIs, a set of images $T = \{t_i\}$ with $t_i \in \mathbb{R}^{A \times B}$ were used, each one associated with a dense pixel-level ground truth mask $M = \{m_i\}$, where $m_i \in \mathbb{R}^{A \times B}$, and an image-level scalar label $Y = \{y_i\}$. 

To emulate global patterns in the images (e.g., tumor lesions), we defined two rectangles within each mask placed at random locations and characterized by their own orientation: one was either vertically or horizontally oriented (\textit{non-tilted}); the other was tilted either 45 or 135 degrees (\textit{tilted}). Each rectangle was associated to a randomly selected MNIST~\cite{lecun1998gradient} digit class. To emulate local patterns (e.g., cells), instances of MNIST digits were placed throughout the images at random locations. The class of these instances was determined by their spatial position, i.e., belonging to a certain rectangle class if placed within the boundaries of a rectangle or otherwise randomly selected. The label of each image was defined by the class of the \textit{tilted} rectangle, with the \textit{non-tilted} rectangle acting as a distraction. See Fig.~\ref{fig:synthetic_data} for an example image. 

Note that, in order to solve this classification task, NIC had to detect the \textit{tilted} rectangle and its class without access to the ground-truth masks. Moreover, the method must combine local and global information, i.e., exploiting the local features that identify digit instances' classes while recognizing their global spatial arrangement to detect the orientation of the rectangle. 

We downsampled MNIST digits to $9 \times 9$ pixels, defining a patch size $P=9$ and stride $S=9$ pixels. WSIs are typically $50000\times50000$ pixels in size, with patch sizes of $128\times128$ pixels covering structures composed of a few cells. We mimicked this image-patch ratio by using an image size of $A=B=3600$ pixels, and inserted \num{25920} digit instances per image (\num{0.2}\% of the total possible locations). Rectangle size randomly ranged from \num{1800} pixels to \num{36} pixels (long side). This reduced image size enabled us to run more thorough experiments than what we could do with histopathological data.

A total of \num{50000} images with balanced labels were created across the 10 digit classes: \num{2500} to generate patches to train the encoders, \num{22500} to train the NIC CNN (with 75\% and 25\% for training and validation), and \num{25000} as an independent test set for the NIC CNN.

\subsubsection{Camelyon16 histopathology dataset}
The \emph{Camelyon16}~\cite{camelyon16} dataset is a publicly available multicenter cohort that consists of 400 sentinel lymph node H\&E WSIs from breast cancer patients. Reference standard exists in two forms: fine-grained annotations of metastatic lesions and image-level labels indicating the presence of tumor metastasis in each slide. Sixty WSIs from the original training set were set aside to train encoders at patch level. The remaining WSIs were combined with the original test set (n=340) to train and evaluate a classification model using image-level labels only.

\subsubsection{TUPAC16 histopathology dataset}
The \emph{TUPAC16}~\cite{veta2018predicting} dataset was used, consisting of 492 H\&E WSIs from invasive breast cancer patients. It is a publicly available cohort with WSIs from The Cancer Genome Atlas~\cite{weinstein2013cancer} where each WSI is associated with a tumor proliferation speed score, an objective measurement that takes into account the RNA expression of 11 proliferation-associated genes~\cite{nielsen2010comparison}. We set aside 40 WSIs from this set to train encoders at patch level. The remaining WSIs (n=452) were used to train and evaluate a regression model using image-level labels only. Additionally, 321 test WSIs with no public ground truth available were used to perform an independent evaluation.

\subsubsection{Rectum histopathology dataset}
The \emph{Rectum} dataset is a publicly available set of 74 H\&E WSIs from rectal carcinoma patients~\cite{ciompi2017importance}. Manual annotations of 9 tissue classes were made by an expert: blood cells, fatty tissue, epithelium, lymphocytes, mucus, muscle, necrosis, stroma, and tumor. The slides were randomized and organized into ten equal partitions at patient level, five of which were used for training, one for validation, and four for testing. This dataset was used to train and evaluate encoders at patch level only. We extracted a balanced distribution of 15K, 852, and 4K patches per class from the training, validation, and test slides, respectively.

\subsubsection{Data preparation}
\label{sec:data_preparation}
Regarding the synthetic dataset, one million pairs of patches were extracted to train the encoders, augmented with scaling and elastic deformation. To avoid creating a dataset of empty patches, the probability of sampling a patch containing a white pixel was twice of that of an empty patch.

All WSIs in this study were preprocessed with a tissue-background segmentation algorithm~\cite{bandi2017comparison} in order to exclude areas not containing tissue from the analysis. Furthermore, all images were analyzed at \SI{0.5}{\um/pixel} resolution.

A set of patch datasets were assembled to train and evaluate each of the encoding networks described in Sec. \ref{sec:wsi_compression} using the set of images that we set aside from each cohort: 60 WSIs from \textit{Camelyon16}, 40 from \textit{TUPAC16}, and all from \textit{Rectum}. Each of these subcohorts were divided into training, validation, and test partitions. 

The \emph{contrastive dataset} was created by extracting an equal amount of patches from each source (i.e., \textit{Camelyon16}, \textit{TUPAC16}, and \textit{Rectum}) and merged into \SI{50000} and \SI{25000} patch pairs for training and validation, respectively. The \emph{non-contrastive dataset} was then created by randomizing all individual patches within the \emph{contrastive dataset}. 

The \emph{supervised-tumor dataset} was created by extracting \SI{50000}, \SI{10000}, and \SI{50000} patches from the set of 60 \textit{Camelyon16} WSIs for training, validation, and testing, respectively. Finally, the \emph{supervised-tissue dataset} consisted of the \textit{Rectum} training, validation, and test sets containing \SI{131000}, \SI{8000}, and \SI{35000} patches, respectively. Note that the patches in the \emph{supervised-tumor dataset} and \emph{supervised-tissue dataset} test sets did not undergo any augmentation. The fine-grained tumor annotations were used to sample a balanced distribution of tumor and non-tumor patches in the \emph{supervised-tumor dataset} and 9-class patches in the \emph{supervised-tissue dataset}.

\begin{figure*} [t]
\begin{center}
\includegraphics[width=1\textwidth]{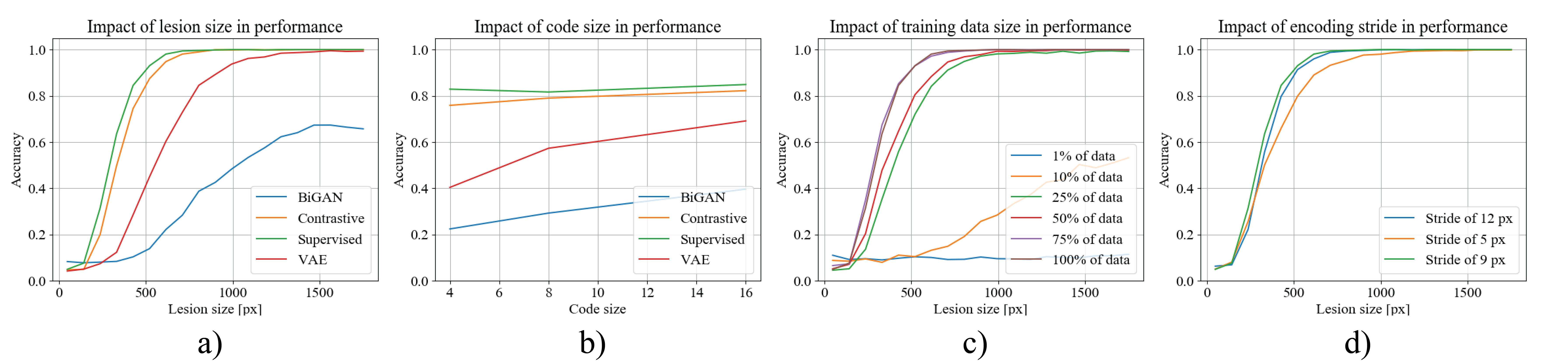}
\end{center}
\caption[syntheticresults] 
{ \label{fig:synthetic_results} Experimental results with synthetic data and image-level labels. Default hyper-parameter choice unless specified otherwise is: \textit{supervised} encoder, code size 16, stride 9 pixels, and usage of 100\% of training data. }
\end{figure*} 

\begin{figure} [t]
\begin{center}
\includegraphics[width=0.5\textwidth]{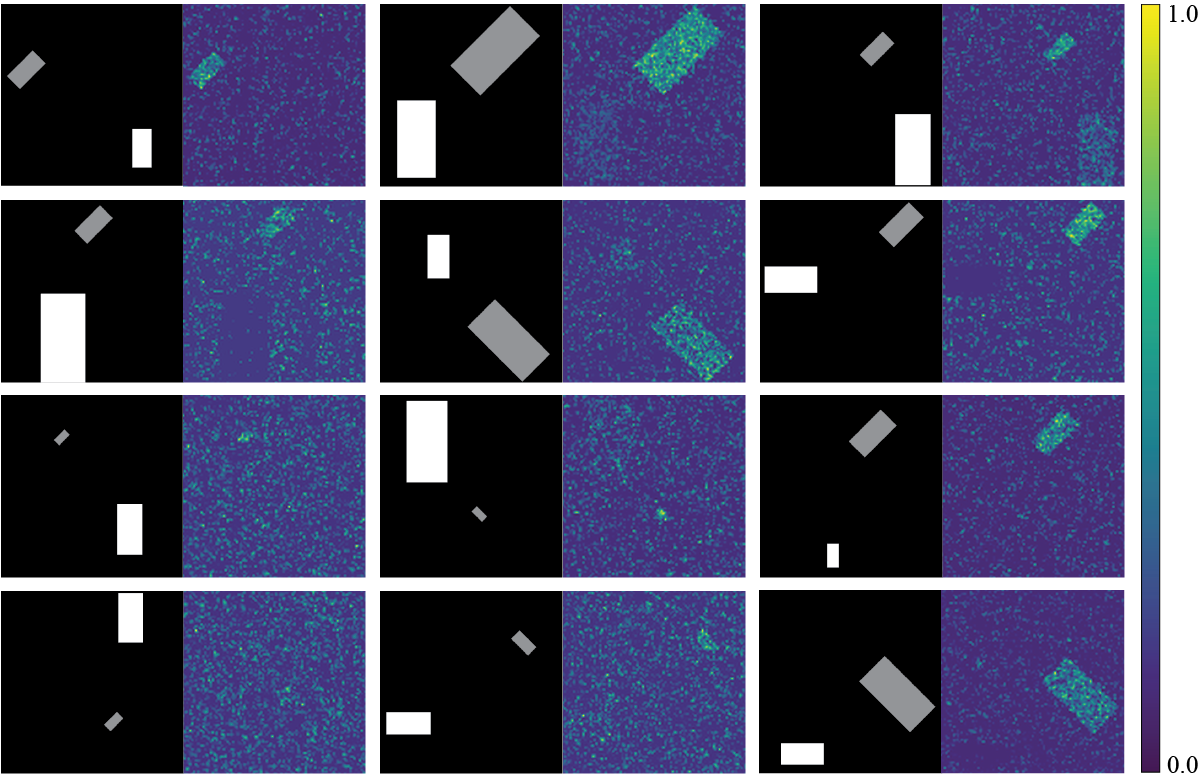}
\end{center}
\caption[syntheticgradcam] 
{ \label{fig:synthetic_gradcam} Grad-CAM visualization applied to randomly selected synthetic test images. Left images within the pairs correspond to the ground truth masks (unseen by the model), and right ones to the saliency heatmaps. Note that areas corresponding to the grey \textit{tilted} rectangles (responsible for the image-level labels) are highly salient with respect to the rest of the image.}
\end{figure} 

\subsection{Experimental results on synthetic data}
\label{sec:syn_task}

The \textit{contrastive} encoder was trained using the pairs of patches described in Sec.~\ref{sec:data_preparation}. The \textit{VAE} and \textit{BiGAN} encoders were subsequently trained using these same patches, concatenating and shuffling them along the pair dimension. Finally, a \textit{supervised} encoder was trained with MNIST digits to serve as an oracle feature extractor. Once the encoders were trained, all images were encoded to produce a different embedded representation for each encoding configuration. Network architectures and training protocols are detailed in the Supplementary Material accompanying this paper. 

We explored different values for the method hyper-parameters (e.g., code size and stride) using the synthetic data, and evaluated the accuracy of each resulting CNN in the independent test set. We analyzed how this performance was affected by the size of the simulated lesion, i.e., the size of the \textit{tilted} rectangle. Results are summarized in Fig.~\ref{fig:synthetic_results}. Overall, the \textit{contrastive} encoder achieved the best performance among the unsupervised techniques, very close to that of the oracle, followed by the \textit{VAE} and \textit{BiGAN} encoders. This trend was maintained when analyzing the impact of the lesion size. We found out that the method's performance degraded quickly when the size of the target lesion was smaller than 10\% of the image size (see Fig.~\ref{fig:synthetic_results}-a).

Additionally, the performance impact of the code size used to compress the images was assessed (Fig.~\ref{fig:synthetic_results}-b). It was observed that larger code sizes generally improved performance, a result that was more evident for less accurate encoding methods like \textit{VAE} and \textit{BiGAN}. Subsequently, different stride values were tested using the oracle encoder and a code size of 16: it was found that a smaller stride, producing embedded images with larger spatial resolution, resulted in hampered performance (Fig.~\ref{fig:synthetic_results}-c). Finally, the impact of training data size in performance was analyzed using the oracle encoder with code size 16 and stride 9 (Fig.~\ref{fig:synthetic_results}-d). These results indicate that NIC required in the order of thousands of images to perform well, a requisite that is rarely met in real histopathological datasets.

In our last experiment, we applied Grad-CAM to visualize the regions of the input images that were responsible for the CNN prediction (see Fig.~\ref{fig:synthetic_gradcam}). Remarkably, the network seemed to be able to discern between background noise and the rectangular patterns. Upon visual inspection, the CNN generally focused on the \textit{tilted} rectangle, the one responsible for the image-level label. We applied a simple general-purpose post-processing routine to denoise the heatmaps and reject spurious activity. We measured the Jaccard similarity coefficient per image between the post-processed heatmap and the ground truth maps, and obtained 0.612 on average across test images.

\begin{table*}[]
\caption{Patch-level classification performance (accuracy). \textit{Task-1} and \textit{Task-2} in the text refer to columns \textit{Camelyon-Tumor} and \textit{Rectum-Global}. Reporting mean and standard deviation using two random weight initializations.}
\label{tab:exp_patch}
\centering
\scriptsize\addtolength{\tabcolsep}{-3pt}
\begin{tabular}{l|l|llllllllll}
\hline
                  & \textbf{Camelyon} & \multicolumn{10}{c}{\textbf{Rectum}}                                                                                                                                     \\ \hline
\textbf{Encoder}  & \textbf{Tumor}    & \textbf{Blood} & \textbf{Fat}   & \textbf{Epith} & \textbf{Lymph} & \textbf{Mucus} & \textbf{Muscle} & \textbf{Necro} & \textbf{Strom} & \textbf{Tumor} & \textbf{Global} \\ \hline

VAE & 0.799(0.004) & 0.602(0.034) & 0.735(0.154) & 0.556(0.006) & 0.811(0.018) & 0.623(0.125) & \textbf{0.823(0.014)} & 0.170(0.018) & \textbf{0.768(0.008)} & 0.667(0.000) & 0.639(0.010) \\
Contrastive & 0.789(0.004) & 0.304(0.018) & \textbf{0.966(0.003)} & 0.502(0.005) & 0.850(0.014) & 0.240(0.011) & 0.609(0.006) & 0.140(0.010) & 0.595(0.005) & 0.476(0.014) & 0.520(0.002) \\
BiGAN & \textbf{0.806(0.022)} & \textbf{0.738(0.034)} & 0.879(0.059) & \textbf{0.627(0.000)} & \textbf{0.899(0.008)} & \textbf{0.802(0.055)} & 0.796(0.002) & \textbf{0.769(0.021)} & 0.601(0.066) & \textbf{0.770(0.010)} & \textbf{0.765(0.013)} \\ \hline
Mean-RGB & 0.772(0.001) & 0.736(0.000) & 0.635(0.355) & 0.202(0.049) & 0.385(0.068) & 0.720(0.270) & 0.904(0.008) & 0.030(0.028) & 0.668(0.039) & 0.252(0.000) & 0.504(0.022) \\
Sup.-tumor & 0.855(0.001) & 0.578(0.090) & 0.896(0.005) & 0.400(0.007) & 0.981(0.004) & 0.868(0.021) & 0.507(0.061) & 0.494(0.049) & 0.467(0.027) & 0.618(0.019) & 0.646(0.008) \\
Sup.-tissue & 0.800(0.006) & 0.835(0.003) & 0.958(0.008) & 0.832(0.029) & 0.935(0.010) & 0.937(0.026) & 0.940(0.002) & 0.906(0.005) & 0.863(0.009) & 0.934(0.002) & 0.904(0.000) \\ \hline

\end{tabular}
\end{table*}

\begin{table}[t]
\caption{Predicting the presence of metastasis at WSI level (AUC). Reporting mean and standard deviation using two random weight initializations.}
\label{tab:exp_c16}
\centering
\begin{tabular}{l|lll}
\hline
\textbf{Encoder}  & \textbf{All} & \textbf{Test} & \textbf{Macro} \\
\hline

VAE & 0.661(0.007) & 0.671(0.008) & 0.634(0.003) \\
Contrastive & 0.608(0.001) & 0.651(0.016) & 0.606(0.012) \\
BiGAN & \textbf{0.725(0.009)} & \textbf{0.704(0.030)} & \textbf{0.720(0.010)} \\ \hline
Mean-RGB & 0.582(0.006) & 0.578(0.016) & 0.585(0.014) \\
Supervised-tumor & 0.760(0.002) & 0.771(0.002) & 0.914(0.000) \\

\hline

\end{tabular}
\end{table}

\subsection{Training of encoders}

Due to the computationally expensive nature of experimenting with gigapixel WSIs, we only tested a subset of the hyper-parameters that we explored with synthetic data. We selected their values using the following heuristics. We used $P=128$, a common patch size used in the Computational Pathology literature~\cite{bandi2017comparison}, with a stride of the same size $S=128$ to perform non-overlapping patch sampling. We selected $R=400$ to obtain crops corresponding to typical sizes of gigapixel WSIs ($\num{50000} \times \num{50000}$ pixels) and $T=10$ as done in the literature~\cite{krizhevsky2012imagenet}. Finally, we selected $C=128$ to perform our experiments using a single GPU. Network architectures and training protocols are detailed in the Supplementary Material.

We trained the \textit{contrastive} encoder with the \emph{contrastive dataset}, and the \textit{VAE} and \textit{BiGAN} models with the \emph{non-contrastive dataset}. Note that these datasets contained the exact same image patches, ensuring a fair comparison among encoders. No manual annotations were required in this process. We trained a supervised baseline encoder for breast tumor classification using the \emph{supervised-tumor dataset}, and a supervised baseline encoder for rectum tissue classification using the \emph{supervised-tissue dataset}.

It is widely recognized that color-based features can be very informative in histopathology image analysis~\cite{sertel2009computer,tabesh2007multifeature,kong2007image}. Therefore, we included an additional encoding function to capture color information from the raw input by averaging the pixel intensity across spatial dimensions from input RGB patches. It provided a simple yet effective baseline to compare with more sophisticated encoding mechanisms.

This entire training process resulted in 6 encoding networks used in subsequent experiments: the \emph{mean-RGB} baseline, \emph{VAE} encoder, \emph{contrastive} encoder, \emph{BiGAN} encoder, \emph{supervised-tumor} baseline, and \emph{supervised-tissue} baseline.

\subsection{Comparing encoding performance}

Due to the lack of a common evaluation methodology for unsupervised representation learning, we compared the performance of these 6 encoders when used as fixed feature extractors for related supervised classification tasks. We defined two tasks: (1) discerning between tumor and non-tumor patches on the \emph{supervised-tumor dataset} (\textit{Task-1}), and (2) performing 9-class tissue classification on the \emph{supervised-tissue dataset} (\textit{Task-2}). For each task, we trained an MLP on top of each encoder with frozen weights and reported the accuracy metric for each test set.

Results in Tab. \ref{tab:exp_patch} highlight several observations. First, \emph{VAE}, \emph{contrastive}, and \emph{BiGAN} performed better than the lower baseline for both \emph{Task 1} and \emph{Task 2}, stressing their ability to describe complex patterns beyond simple features related to color intensity. Second, the \emph{VAE} encoder obtained a higher performance than the \emph{contrastive} one, particularly for \emph{Task 2}. Third, the \emph{BiGAN} encoder achieved the best performance among all the unsupervised methods, with a relatively large margin for the more complex \emph{Task 2} with respect to the runner-up \emph{VAE} model. Furthermore, the \emph{BiGAN} encoder obtained the best result for 5 out of 9 classes in \emph{Task 2}, and it achieved the first or second best result for 8 out of 9 classes among the unsupervised models. Remarkably, \emph{BiGAN} succeeded at classifying patches from challenging tissue classes such as blood cells and necrotic tissue.

\begin{table}[t]
\caption{Predicting tumor proliferation speed at WSI level (Spearman corr.). Reporting mean and standard deviation using two random weight initializations.}
\label{tab:exp_t16}
\centering
\begin{tabular}{l|ll}
\hline
\textbf{Encoder}  & \textbf{All} & \textbf{Test} \\
\hline

VAE & 0.419(0.004) & - \\
Contrastive & 0.390(0.006) & - \\
BiGAN & \textbf{0.522(0.001)} & 0.558(0.001) \\ \hline
Mean-RGB & 0.238(0.020) & - \\
Supervised-tumor & 0.427(0.014) & - \\

\hline
\end{tabular}
\end{table}

\subsection{Predicting the presence of metastasis at image level}

In this experiment, we trained a CNN to perform binary classification on compressed gigapixel WSIs from the \textit{Camelyon16} cohort, identifying the presence of tumor metastasis using image-level labels only. Due to the limited amount of images in this cohort (340 WSIs), we divided the dataset into four equal-sized partitions and performed four rounds of cross-validation using two partitions for training, one for validation and one for testing, rotating them in each round. We trained a different CNN classifier for each encoder, i.e., \emph{mean-RGB}, \emph{VAE}, \emph{contrastive}, \emph{BiGAN}, and the upper baseline \emph{supervised-tumor}. We reported the area under the receiver operating characteristic (AUC) on three evaluation sets.

The first evaluation set (\emph{All}) concatenated all samples in each of the hold-out partitions. Note that each hold-out partition was evaluated by a different CNN that had never seen the data. The second evaluation set (\emph{Test}) was a subset of \emph{All} that matched the official test set of the \textit{Camelyon16} Challenge, used for comparison with the public leaderboard. The third evaluation set (\emph{Macro}) used the same WSIs as in \emph{Test} but considering only those that presented a macro metastasis as positive labels, i.e., a tumor lesion larger than \SI{2}{mm}. The macro labels were only available for the \textit{Camelyon16} test set. The \emph{Macro} set was relevant to evaluate how the method performed with lesions visible at low resolution.

Results in Tab. \ref{tab:exp_c16} demonstrate that the method presented in this work is an effective technique for gigapixel image analysis using image-level labels only. Regarding the \emph{All} evaluation set, \emph{BiGAN} achieved a remarkable performance of 0.716 AUC, with a relative difference from the \textit{supervised} baseline of only 6\%. The \emph{contrastive} and \emph{VAE} models also surpassed the lower baseline, but obtained substantially lower performance scores compared to \emph{BiGAN}. Regarding the \emph{Test} set, the \emph{BiGAN} encoder obtained a lower performance of 0.674 AUC. In the \emph{Macro} set, the performance gap between the \textit{supervised} baseline and the \emph{BiGAN} encoder increased substantially from 0.095 to 0.184. The state-of-the-art in \textit{Camelyon16} obtained 0.9935 AUC in the \textit{Test} set using accurate pixel-level annotations to train their model.

\begin{figure} [t]
\centering
\includegraphics[width=0.4\textwidth]{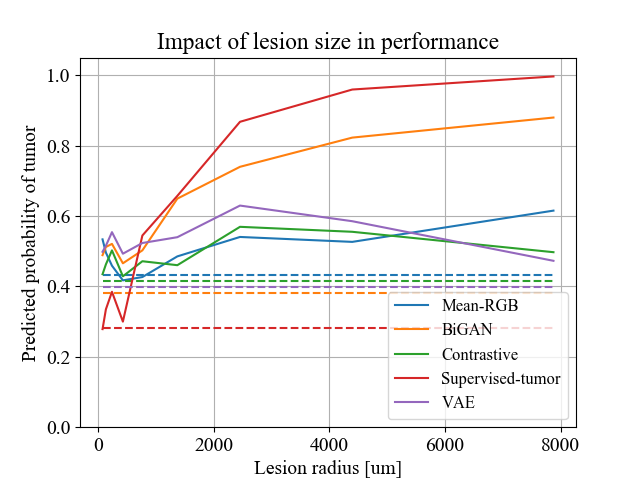}
\caption[problesionsize] 
{ \label{fig:plot_lesion_size} Experimental results with respect to lesion size in \textit{Camelyon16} \textit{all} test set using multiple encoders. Solid lines: average probability of samples with positive labels; dashed lines: average probability of samples with negative labels (no lesion).}
\end{figure} 

\begin{figure} [t]
\centering
\includegraphics[width=0.4\textwidth]{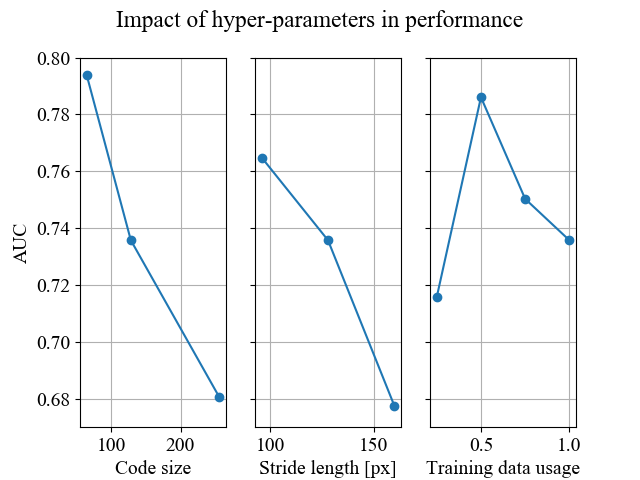}
\caption[codestride] 
{ \label{fig:plot_code_stride} Hyper-parameter value analysis performed in \textit{Camelyon16} data using the \textit{supervised} encoder. Evaluated on unseen images from the first data partition out of the 4-fold cross-validation sets. Left: varying code size using a fix stride of 128 pixels; center: varying stride while using a fix code size of 128 elements; and right: varying the number of WSIs used during training. }
\end{figure}

Additionally, we analyzed the performance of our method as a function of the lesion size in the \textit{All} test set. The lesion size is a measurement determined by pathologists taking the distribution of tumor cell clusters within a WSI into account. Since this annotation was not available for all WSIs, we approximated it by computing the radius of an hypothetical circle with an area composed of all pixels annotated as tumor in each WSI. Results in Fig.~\ref{fig:plot_lesion_size} indicated that our method's performance degraded with small tumor lesions across most encoders, in line with the results obtained with synthetic data. Furthermore, we experimented with different hyper-parameters such as code size, stride, and training data size using the \textit{supervised} encoder (Fig.~\ref{fig:plot_code_stride}). We found that performance improvements might be gained from careful hyper-parameter tuning of the code size and stride parameters. Moreover, there seemed to be a weak but positive correlation between model performance and training data size.

\subsection{Predicting tumor proliferation speed at image level}

In this experiment, we trained a CNN to perform a regression task on compressed gigapixel WSIs from the \textit{TUPAC16} cohort, predicting the tumor proliferation speed based on gene-expression profiling. We performed 4-fold cross-validation as in the previous experiment, and reported the Spearman correlation between the predicted and the true scores of two evaluation sets. 

The first evaluation set (\emph{All}) concatenated all samples in each of the hold-out partitions. The second evaluation set (\emph{Test}) matched the test set used in the \textit{TUPAC16} Challenge, whose ground truth is not public. Using the encoder that obtained the highest performance, we evaluated each WSI in \emph{Test} four times using each of the CNNs trained during cross-validation and submitted the average score per slide. Our predictions were independently evaluated by the challenge organizers, ensuring a fair and independent comparison with the state of the art.

The results in Tab. \ref{tab:exp_t16} showed that \emph{BiGAN} achieved the highest performance with a 0.521 Spearman correlation. Remarkably, this score was superior to that of any other unsupervised or supervised encoder. In addition, we obtained a score of 0.557 on the \textit{TUPAC16} Challenge test set, superior to the state-of-the-art for image-level regression with a score of 0.516. Note that the first entry of the leaderboard used an additional set of manual annotations of mitotic figures, thus it cannot be compared with our setup. 

\begin{figure*} [t]
\begin{center}
\includegraphics[width=1\textwidth]{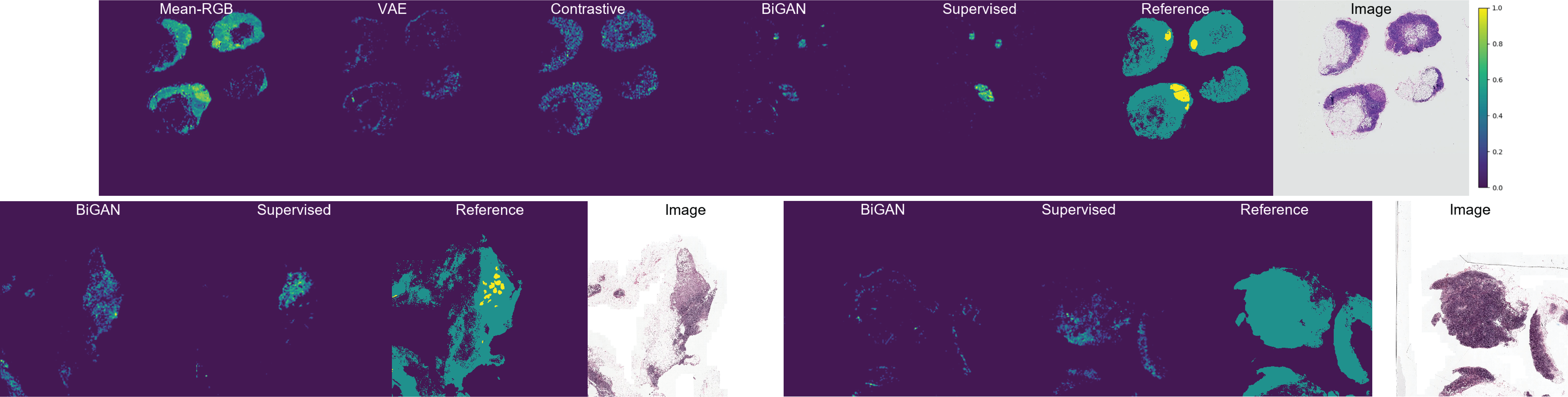}
\end{center}
\caption[gradcam_c16] 
{ \label{fig:gradcam_c16} Grad-CAM visualization applied to several WSIs from \textit{Camelyon16}. Top: the first five images represent the saliency maps for CNNs trained with 5 different encoders, respectively. The sixth and seventh images are the reference standard (manual annotations) and RGB thumbnail of the WSI, respectively. Dark blue represents low saliency, whereas yellow indicates high saliency. Bottom-left: failure case where the \textit{BiGAN} model failed to recognize the tumor area. Bottom-right: failure case where the \textit{BiGAN} model attended to a region with no tumor cells.}
\end{figure*} 

\begin{figure*} [t]
\begin{center}
\includegraphics[width=0.85\textwidth]{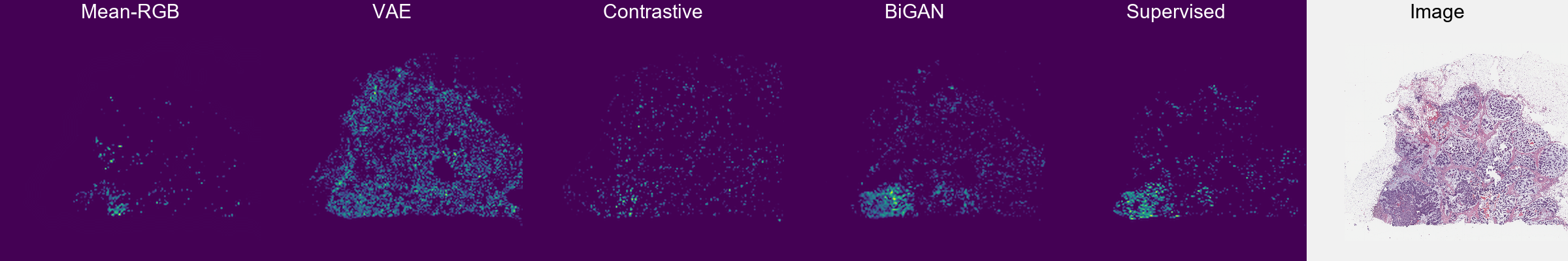}
\end{center}
\caption[gradcam_t16] 
{ \label{fig:gradcam_t16} Grad-CAM visualization applied to a sample case from \textit{TUPAC16}. The first five images represent the saliency maps for CNNs trained with five different encoders, respectively. The last image is an RGB thumbnail of the WSI. Dark blue represents low saliency, whereas yellow indicates high saliency. }
\end{figure*} 

\subsection{Visualizing \emph{where} the information is located}

We conducted a qualitative analysis on the trained CNNs to locate the spatial position of visual cues relevant in predicting the image-level labels. We applied the Grad-CAM algorithm to the CNNs trained for both tasks at image level. For the tumor metastasis prediction task, we compared the saliency maps with fine-grained manual annotations. Figures~\ref{fig:gradcam_c16} and~\ref{fig:gradcam_t16} include the results for a few samples; the results for the remaining WSIs can be found in the Supplementary Material. Note that each WSI was evaluated by a CNN that had not yet seen the image (hold-out partition).

Fig.~\ref{fig:gradcam_c16} shows that the \emph{mean-RGB} baseline model lacked the ability to focus on specific tissue regions, suggesting that it was unable to learn discriminative features from image-level labels. The \emph{VAE} and \emph{contrastive} models exhibited a suboptimal behavior, scattering attention all over the image. Remarkably, the \emph{BiGAN} model seemed to focus on tumor regions only, discarding empty areas, fatty tissue, and healthy dense tissue. It showed a strong discriminative power to discern between tumor and non-tumor regions, even though the CNN had access to image-level labels only. For completeness, we also included the \emph{supervised-tumor} baseline that also exhibited a focus on tumor regions. Nevertheless, these heatmaps are often difficult to interpret and cannot be used for a more quantitative analysis. Failure cases can be seen in the bottom part of Fig.~\ref{fig:gradcam_c16}, where the CNN highlighted non-tumorous regions.

Regarding Fig.~\ref{fig:gradcam_t16}, a similar trend to the one found in the previous task was observed for all encoders: the \emph{BiGAN} model focused on very specific regions of the WSIs that seemed compatible with active tumor regions. The \emph{supervised-tumor} baseline focused on irrelevant areas, in line with its poor performance for this task.

\section{Discussion}
\label{sec:discussion}

Our experimental results support the hypothesis that visual cues associated with weak image-level labels can be exploited by our method, integrating information from global structure and local high-resolution visual cues. Furthermore, we have shown that this methodology is flexible and completely label-agnostic, delivering relevant results for both classification and regression tasks in synthetic as well as histopathological data. It emerges as a promising strategy to tackle the analysis of more challenging image-level labels that are closely related to patient outcome, e.g., overall survival and recurrence-free survival. Gigapixel NIC paves the way for leveraging existing computer vision algorithms that could not be applied in the gigapixel domain until now, such as image captioning (useful to generate written clinical reports), visual question answering, image retrieval (to find similar pathologies), anomaly detection, and generative modeling~\cite{shin2016learning, ImageCLEFVQA-Med2018, anavi2015comparative, schlegl2017unsupervised, yang2018low}.

A key assumption in our method was that high-resolution image patches could be represented by low-dimensional highly compressed embedding vectors. We analyzed several unsupervised strategies to achieve such a compression and found that the \emph{BiGAN} encoder, trained using adversarial feature learning, was superior to all other methods across all experiments with histopathological data. We believe that this relative improvement with respect to the \emph{VAE} and \emph{contrastive} methods is explained by intrinsic algorithmic differences among the methods. In particular, the \emph{VAE} model relies on minimizing the MSE objective, which is a unimodal function that fails to capture high-level semantics; it focuses on reconstructing low-level pixel information instead, wasting embedding capacity. On the other hand, the \emph{contrastive} encoder uses the embedding capacity more efficiently, but its performance is driven by the design of the hand-engineered contrastive task. Remarkably, the \emph{BiGAN} model learns an encoder that fully automatically inverts a complex mapping between the latent space and the image space. By doing so, the encoder benefits from all the high-level features and semantics already discovered by the generator, producing very effective discriminative embedding vectors. Furthermore, \textit{BiGAN} achieved the best classification accuracy on the challenging blood, mucus, and necrotic tissue classes that rarely appear in the \textit{Camelyon16} and \textit{TUPAC16} WSIs. We hypothesize that the adversarial method can model these rare data modes more effectively than the \emph{contrastive} or \emph{VAE} approaches. Nevertheless, we believe that the choice of encoder may be data-dependent, since the \textit{contrastive} encoder outperformed the other approaches in the synthetic dataset.

We trained a CNN to predict the breast tumor proliferation speed based on gene-expression profiling, a label associated with unknown visual cues. Our method succeeded in finding and exploiting these patterns in order to predict expected tumor proliferation speed, surpassing the current state-of-the-art for image-level based methods. This shows that our method constitutes an effective solution to deal with gigapixel image-level labels with unknown associated visual cues. Moreover, our method could be used in future works to effectively mine datasets with thousands of gigapixel images~\cite{coudray2018classification}; other automatically generated labels from immunohistochemistry, genomics, or proteomics can be targeted, and visual patterns beyond the knowledge of human pathologists may be discovered.

For the first time, the regions of a gigapixel image that a trained CNN attends to when predicting image-level labels were visualized, and the effect of different encoding methods was compared. We discovered that only the CNNs trained with images compressed with the \emph{BiGAN} encoder and the \emph{supervised-tumor} baseline were able to attend to regions of the image where tumor cells were present. The fact that the \emph{BiGAN} model simultaneously learned to delimit metastatic lesions and identify tumor features within the patch embeddings validates our hypothesis that CNNs are an effective method for analyzing gigapixel images, i.e., since they can exploit both global and local context.

We targeted the presence of tumor metastasis in breast lymph nodes and showed that the \emph{BiGAN} setup performed similarly to the \textit{supervised} baseline. However, our best-performing algorithm was still inferior to that of the \textit{Camelyon16} leadingboard (0.9935 AUC using accurate pixel-level annotations). This performance gap is likely due to two factors. First, the majority of the images marked as positive contain tumor lesions comprised of only a few tumor cells (i.e., micro-metastasis), becoming almost undetectable with the compression setup tested in this work (see Fig.~\ref{fig:plot_lesion_size}). Second, the lack of training data (only a few hundred training images) may lead the CNN into the overfitting regime.

We acknowledge several limitations of our method. For one, it requires a substantial amount of I/O throughput and storage due to the need to write compressed WSI representations to disk before training, and repetitively read them to assemble mini-batches during training. This computational burden prevented us from performing a wide hyper-parameter value search, which may have resulted in a suboptimal parameter selection. Second, it was also observed that the method's performance was proportional to lesion size. In particular, it struggled to detect micro-metastasis in \textit{Camelyon16} data, i.e., tumor lesions smaller than \SI{2}{mm}, limiting the applicability of NIC to tasks with large lesions.

This method can be extended in multiple ways. More sophisticated encoders may improve the low-dimensional representation of the image patches~\cite{oord2018representation,bang2018high,caron2018deep}. Incorporating attention mechanisms may make it easier for the CNN to attend to relevant regions for the image-level labels~\cite{jetley2018learn}, improving the detection of small lesions. Finally, gradient checkpointing~\cite{chen2016training} could be used to backpropagate the training signal from the image-level labels towards the encoder weights.

\section{Conclusion}
\label{sec:conclusion}

Our method for gigapixel neural image compression was able to distill relevant information into compact image representations. The fact that a CNN could be trained using these alternative learned representations opens opportunities to use other methods: gigapixel images are no longer considered as low-level pixel arrays, but operate in a higher level of abstraction. In this work, we showed examples of classification, regression, and visualization performed in a latent space learned by a neural network. These positive results enable performing more advanced gigapixel applications in the latent space, such as data augmentation, generative modeling, content retrieval, anomaly detection, and image captioning.


%

\section*{Acknowledgment}

This study was supported by a Junior Researcher grant from the Radboud Institute of Health Sciences (RIHS), Nijmegen, The Netherlands; a grant from the Dutch Cancer Society (KUN 2015-7970); and another grant from the Dutch Cancer Society and the Alpe d'HuZes fund (KUN 2014-7032); this project has also been partially funded by the European Union's Horizon 2020 research and innovation programme under grant agreement No 825292. The authors would like to thank Dr. Mitko Veta for evaluating our predictions in the test set of the \textit{TUPAC16} dataset, and the developers of Keras~\cite{chollet2015keras}, the open source tool that we used to run our deep learning experiments.

\ifCLASSOPTIONcaptionsoff
  \newpage
\fi



\bibliographystyle{IEEEtran}
\bibliography{mybib}


\vskip -1.5\baselineskip plus -1fil

\begin{IEEEbiography}[{\includegraphics[width=1in,height=1.25in,clip,keepaspectratio]{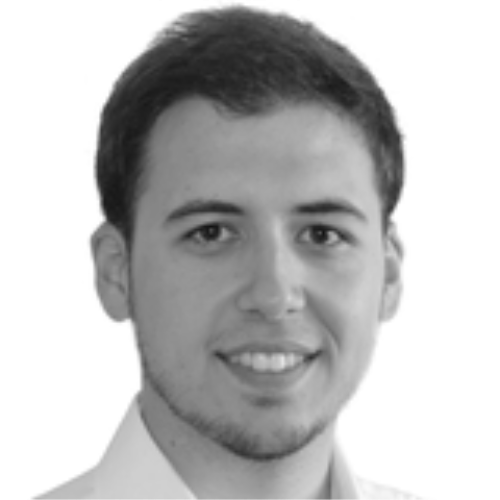}}]{David Tellez} received the BSc and MSc degrees in telecommunication engineering from the University of Seville, Spain, in 2014. He is a PhD candidate in the Computational Pathology Group of Jeroen van der Laak in the Radboud University Medical Center, The Netherlands. His research interests include computer vision and medical imaging.
\end{IEEEbiography}

\vskip -1.5\baselineskip plus -1fil

\begin{IEEEbiography}[{\includegraphics[width=1in,height=1.25in,clip,keepaspectratio]{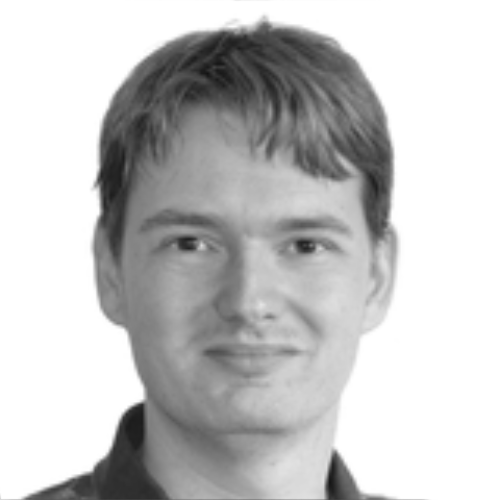}}]{Geert Litjens} completed his PhD thesis on "Computerized detection of prostate cancer in multi-parametric MRI" at the Radboud University Medical Center. He is currently assistant professor in Computation Pathology at the same institution. His research focuses on applications of machine learning to improve oncology diagnostics.
\end{IEEEbiography}

\vskip -1.5\baselineskip plus -1fil

\begin{IEEEbiography}[{\includegraphics[width=1in,height=1.25in,clip,keepaspectratio]{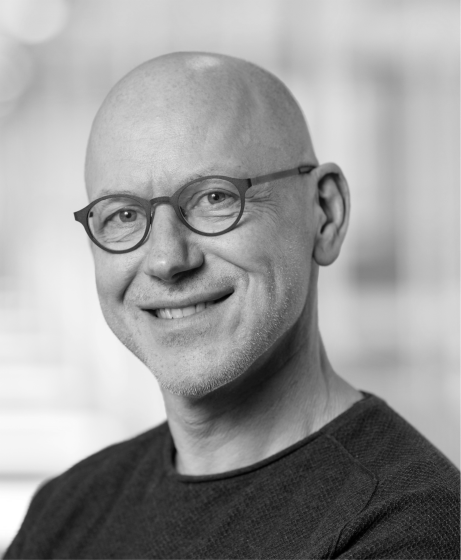}}]{Jeroen van der Laak} holds an MSc in computer science and acquired his PhD from the Radboud University Medical Center in Nijmegen, The Netherlands, where he currently is associate professor of Computational Pathology. His research focuses on deep learning algorithm development and validation for improved pathology diagnostics.​
\end{IEEEbiography}

\vskip -1.5\baselineskip plus -1fil

\begin{IEEEbiography}[{\includegraphics[width=1in,height=1.25in,clip,keepaspectratio]{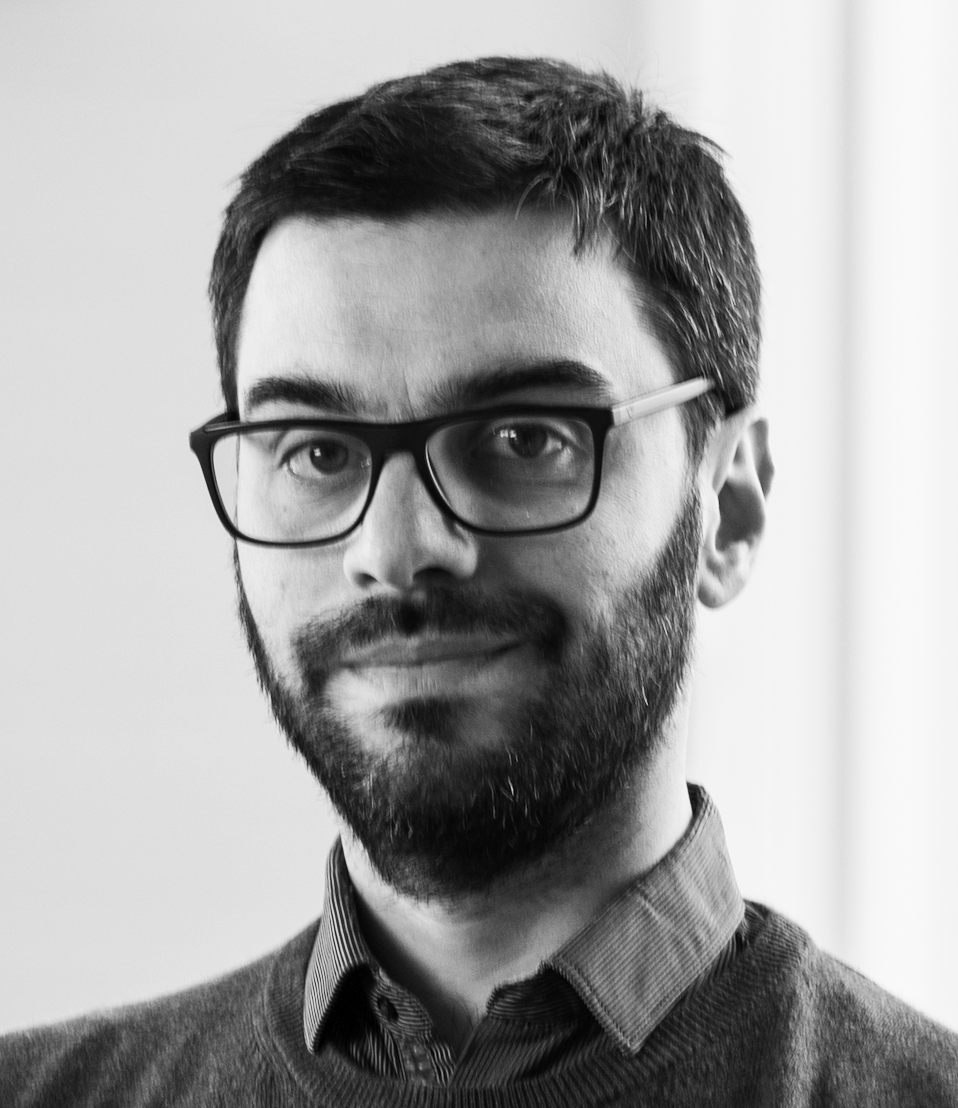}}]{Francesco Ciompi} received the MSc degree in Computer Vision and Artificial Intelligence from the Autonomous University of Barcelona in 2008. In July 2012 he obtained the PhD Cum Laude from the University of Barcelona. Since 2013, he is part of the Diagnostic Image Analysis Group of Radboud University Medical Center. He is Assistant Professor of Computational Pathology, working on Deep Learning for automatic analysis of digital pathology whole-slide images. 
\end{IEEEbiography} 







\newpage

\appendix

\section*{Encoders for Histopathological Data}

\subsection*{Variational Autoencoder}

Two networks are trained simultaneously, the encoder $E$ and the decoder $D$. The task of $E$ is to map an input patch $x~\in~\mathbb{R}^{P\times P\times 3}$ to a compact embedded representation $e\in~\mathbb{R}^{C}$, and the task of $D$ is to reconstruct $x$ from $e$, producing $x'~\in~\mathbb{R}^{P\times P\times 3}$. In this work, we used a more sophisticated version of AE, the variational autoencoder (VAE)~\cite{kingma2013auto}, with $P=128$ and $C=128$. 

The encoder in the VAE model learns to describe $x$ with an entire probability distribution, in particular, given an input $x$, the encoder $E$ outputs $\mu\in~\mathbb{R}^{C}$ and $\sigma\in~\mathbb{R}^{C}$, two embeddings representing the mean and standard deviation of a normal distribution so that:

\begin{equation}
\label{eq:vae}
e = \mu + \sigma \odot n
\end{equation}

with $n \sim \mathcal{N}(0, 1)$ and $\odot$ denoting element-wise multiplication. 

The architecture of $E$ consisted of 5 layers of strided convolutions with 32, 64, 128, 256 and 512 $3\times3$ filters, batch normalization (BN) and leaky-ReLU activation (LRA); followed by a dense layer with 512 units, BN and LRA; and a linear dense layer with $C$ units. 

The architecture of the decoder $D$ consisted of a dense layer with 8192 units, BN and LRA, eventually reshaped to $(4\times 4\times 512)$; followed by 5 upsampling layers, each composed of a pair of nearest-neighbor upsampling and a convolutional operation~\cite{odena2016deconvolution}, with 256, 128, 64, 32 and 16 $3\times3$ filters, BN and LRA; finalized with a convolutional layer with 3 $3\times3$ filters and tanh activation. 

We trained the VAE model by optimizing the following objective:

\begin{multline}
\label{eq:vae_objective}
\mathcal{V}_{\text{VAE}}(x, n, \theta_E, \theta_D) = \\
= \underset{E, D}{\min{}} \Big[ \underbrace{ \big(x - D(E(x, n))\big)^2}_\text{Reconstruction error} + \underbrace{ \gamma (1 + \log{\sigma^2} - \mu^2 - \sigma^2)}_\text{KL divergence} \Big]
\end{multline}

with $x$ representing a single data sample, $n$ a sample from $\mathcal{N}(0, 1)$, $\gamma$ a scaling factor, and $\theta_E$ and $\theta_D$ as the parameters of $E$ and $D$, respectively. Note that we optimized $\theta_E$ and $\theta_D$ to minimize both the reconstruction error between the input and output data distributions, and the KL divergence between the embedding distribution and the normal $\mathcal{N}(0, 1)$ distribution with $\gamma = \num{5e-5}$. 

We minimized $\mathcal{V}_{\text{VAE}}$ using stochastic gradient descent with Adam optimization and 64-sample mini-batch, decreasing the learning rate by a factor of 10 starting from \SI{1e-3} every time the validation loss plateaued until \SI{1e-5}{}. Finally, we selected the encoder $E$ corresponding to the VAE model with the lowest validation loss.

\subsection*{Contrastive Training}

We assembled a training dataset composed of pairs of patches $\boldsymbol{x} = \{x^{(1)}, x^{(2)}\}$ with $x^{(i)}~\in~\mathbb{R}^{P\times P\times 3}$ where each pair $\boldsymbol{x}$ was associated with a binary label $y$, and $P=128$. In order to solve this binary classification task, we trained a two-branch Siamese network~\cite{melekhov2016siamese} called $S$. Both input branches shared weights and consisted of the same encoding architecture $E$ as the VAE model. After concatenation of the resulting embedding vectors, a MLP followed consisting of a dense layer with 256 units, BN and LRA; finalized by a single sigmoid unit.

We minimized the binary cross-entropy loss using stochastic gradient descent with Adam optimization and 64-sample mini-batch, decreasing the learning rate by a factor of 10 starting from \SI{1e-2} every time the validation classification accuracy plateaued until \SI{1e-5}{}. Finally, we selected the encoder $E$ corresponding to the $S$ with the highest validation classification accuracy.

\subsection*{Bidirectional Generative Adversarial Network}

We trained a BiGAN setup consisting of three networks: a generator $G$, a discriminator $D$ and an encoder $E$. $G$ mapped a latent variable $z$ drawn from a normal distribution $\mathcal{N}(0, 1)$ into artificial images $x'$:

\begin{equation}
\label{eq:bigan_g}
z~\sim \mathcal{N}(0, 1) \in~\mathbb{R}^{C}  \xrightarrow{G} x'~\in~\mathbb{R}^{P\times P\times 3}
\end{equation}

whereas $E$ mapped images $x$ sampled from the true data distribution $\mathcal{X}$ into embeddings $e$:

\begin{equation}
\label{eq:bigan_e}
x~\sim \mathcal{X} ~\in~\mathbb{R}^{P\times P\times 3} \xrightarrow{E} e \in~\mathbb{R}^{C} 
\end{equation}

During training, the three networks played a minimax game where the discriminator $D$ tried to distinguish between \emph{actual} and \emph{artificial} image-embedding pairs, i.e. $\{x, e\}$ and $\{x', z\}$ respectively, while $G$ and $E$ tried to fool $D$ by producing increasingly more realistic images $x'$ and embeddings $e$, i.e. closer to $\mathcal{N}(0, 1)$. We used $P=128$ and $C=128$.

Given the difficulty of training a stable BiGAN model, we downsampled $x$ by a factor of 2 before feeding it to the model. The architecture of the encoder $E$ consisted of 4 layers of strided convolutions with 128 $3\times3$ filters, BN and LRA; followed by a linear dense layer with $C$ units. 

The architecture of the generator $G$ consisted of a dense layer with 1024 units, BN and LRA, eventually reshaped to $(4\times 4\times 64)$; followed by 4 upsampling layers, each composed of a pair of nearest-neighbor upsampling and a convolutional operation~\cite{odena2016deconvolution}, with 128 $3\times3$ filters, BN and LRA; finalized with a convolutional layer with 3 $3\times3$ filters and tanh activation. 

The discriminator $D$ had two inputs, a low-dimensional vector and an image. The image was fed through a network with an architecture equal to $E$ but different weights, and the resulting embedding vector concatenated to the input latent variable. This concatenation layer was followed by two dense layers with 1024 units, LRA and dropout (0.5 factor); finalized with a sigmoid unit. 

We optimized the following objective function:

\begin{multline}
\label{eq:bigan_objective}
\mathcal{V}_{\text{BiGAN}}(x, z, \theta_G, \theta_E, \theta_D) = \\
=\underset{G, E}{\min{}}\underset{D}{\max{}} \Big[\log{\big[ D\big(x, \underbrace{E(x)}_{e}\big) \big]} + \log{\big[ 1 - D\big(\underbrace{G(z)}_{x'},z \big) \big]} \Big]
\end{multline}

with $\theta_G$, $\theta_E$ and $\theta_D$ representing the parameters of $G$, $E$ and $D$, respectively. 

We minimized $\mathcal{V}_{\text{BiGAN}}$ using stochastic gradient descent with Adam optimization, 64-sample mini-batch, and fixed learning rate of \SI{2e-4} for a total of \SI{200000} epochs. Finally, we selected the encoder $E$ corresponding to the last epoch.

\subsection*{Mean-RGB Baseline}

We extracted the embedding $e$ by averaging the pixel intensity across spatial dimensions from input RGB patches $x~\in~\mathbb{R}^{P\times P\times 3}$:

\begin{equation}
\label{eq:bigan_e}
e^{(c)} = \frac{1}{P^2} \sum_{j=1}^{P} \sum_{k=1}^{P} x^{(j,k,c)}
\end{equation} 

with $c$ indexing the three RGB color channels, and $j$ and $k$ indexing the two spatial dimensions.

\subsection*{Supervised-tumor Baseline}

We trained an encoder $E$ identical to the one used in the VAE model, followed by a dense layer with 256 units, BN and LRA; and finalized by a single sigmoid unit.

We minimized the binary cross-entropy loss using stochastic gradient descent with Adam optimization and 64-sample mini-batch, decreasing the learning rate by a factor of 10 starting from \SI{1e-2} every time the validation classification accuracy plateaued until \SI{1e-5}{}. Finally, we selected the encoder $E$ corresponding to the model with the highest validation classification accuracy.

\subsection*{Supervised-tissue Baseline} 

We trained an encoder $E$ identical to the one used in the VAE model, followed by a dense layer with 256 units, BN and LRA; and finalized by a softmax layer with nine units.

We minimized the categorical cross-entropy loss using stochastic gradient descent with Adam optimization and 64-sample mini-batch, decreasing the learning rate by a factor of 10 starting from \SI{1e-2} every time the validation classification accuracy plateaued until \SI{1e-5}{}. Finally, we selected the encoder $E$ corresponding to the model with the highest validation classification accuracy.

\section*{Encoders for Synthetic Data}

We modified the network architectures to account for the smaller patch size selected in the synthetic dataset. We used grayscale patches $x~\in~\mathbb{R}^{P\times P\times 1}$ with $P=9$ and $C=16$ unless stated otherwise. Note that we did not modify the training protocol, e.g. the learning rate schedule.

The architecture of the encoder $E$ used in \textit{VAE}, \textit{Contrastive} and \textit{Supervised} consisted of 2 layers of strided convolutions with 32 and 64 $3\times3$ filters, BN and LRA; followed by a dense layer with 64 units, BN and LRA; and a linear dense layer with $C$ units. 

The architecture of the decoder $D$ used in \textit{VAE} consisted of a dense layer with 2048 units, BN and LRA, eventually reshaped to $(4\times 4\times 128)$; followed by 2 upsampling layers, each composed of a pair of nearest-neighbor upsampling and a convolutional operation~\cite{odena2016deconvolution}, with 64 and 32 $3\times3$ filters, BN and LRA; finalized with a convolutional layer with 1 $3\times3$ filters and tanh activation. 

With respect to \textit{BiGAN}, the architecture of the encoder $E$ consisted of 2 layers of strided convolutions with 32 $3\times3$ filters, BN and LRA; followed by a linear dense layer with $C$ units. The architecture of the generator $G$ consisted of a dense layer with 512 units, BN and LRA, eventually reshaped to $(4\times 4\times 32)$; followed by 2 upsampling layers, each composed of a pair of nearest-neighbor upsampling and a convolutional operation~\cite{odena2016deconvolution}, with 32 $3\times3$ filters, BN and LRA; finalized with a convolutional layer with 1 $3\times3$ filters and tanh activation. The discriminator $D$ had two inputs, a low-dimensional vector and an image. The image was fed through a network with an architecture equal to $E$ but different weights, and the resulting embedding vector concatenated to the input latent variable. This concatenation layer was followed by two dense layers with 128 units, LRA and dropout (0.5 factor); finalized with a sigmoid unit.

\section*{Experiments}

\subsection*{Patch-level Classification}

On top of each encoder with frozen weights, we trained an MLP consisting of a dense layer with 256 units, BN and LRA; followed by either a single sigmoid unit or a softmax layer with nine units, respectively for each classification task. 

We minimized the cross-entropy loss using stochastic gradient descent with Adam optimization and 64-sample mini-batch, decreasing the learning rate by a factor of 10 starting from \SI{1e-2} every time the validation classification accuracy plateaued until \SI{1e-5}{}. Finally, we selected the model with the highest validation classification accuracy.

\subsection*{Image-level Classification and Regression}

We designed a CNN architecture consisting of 8 layers of strided depthwise separable convolutions~\cite{chollet2016xception} with 128 $3\times3$ filters, BN, LRA, feature-wise 20\% dropout, L2 regularization with \SI{1e-5} coefficient, and stride of 2 except for the 7-th and 8-th layers with no stride; followed by a dense layer with 128 units, BN and LRA; and a final output unit, with linear or sigmoid activation for regression or classification tasks, respectively. 

We trained the CNN using stochastic gradient descent with Adam optimization and 16-sample mini-batch, decreasing the learning rate by a factor of 10 starting from \SI{1e-2} every time the validation metric plateaued until \SI{1e-5}{}. We minimized MSE for regression, and maximized binary cross-entropy for binary classification.

\subsection*{Visualizing where the information is located}

Given a compressed gigapixel image $\omega'$, its associated image-level label $y$ and a trained CNN $S$, we performed a forward pass over $\omega'$, producing a set of $J$ intermediate three-dimensional feature volumes $f_j^{(k)}$, with $j$ and $k$ indicating the $j$-th and $k$-th convolutional layer and feature map, respectively. Additionally, we computed the gradients of the feature volume $f_j^{(k)}$ with respect to the class output $y$, for a fixed convolutional layer. We averaged the gradients across the spatial dimensions and obtained a set of gradient coefficients $\gamma_j^{(k)}$ indicating how relevant each feature map was for the desired output $y$. Finally, we performed a weighted sum of the feature maps $f_j^{(k)}$ using the gradient coefficients $\gamma_j^{(k)}$:

\begin{equation}
\label{eq:bigan_e}
h^{(k)} = \sum_{j=1}^{J} f_j^{(k)} \gamma_j^{(k)}
\end{equation}

What we obtained was a two-dimensional heatmap $h^{(k)}$ that highlighted the regions of $\omega'$ that were more relevant for $S$ to predict $y$. In order to maximize the resolution of the generated heatmap, we selected $k=1$. 

Grad-CAM heatmaps for Camelyon16 and TUPAC16 can be found here: https://drive.google.com/drive/folders/16E-06rFbGab6-pXfjpo9vXjBVyUQKUuc

\end{document}